\documentclass{article}

\usepackage{microtype}
\usepackage{graphicx}
\usepackage{subfigure}
\usepackage{booktabs} 

\usepackage{hyperref}

\usepackage[skip=0pt]{caption}
\usepackage{amssymb}
\usepackage{amsmath}
\usepackage{float}
\usepackage{placeins}

\usepackage[accepted]{icml2020}

\icmltitlerunning{Topology Distance}

\begin{document}

\twocolumn[
\icmltitle{Topology Distance: A Topology-Based Approach \\ For Evaluating Generative Adversarial Networks}




\begin{icmlauthorlist}
\icmlauthor{Danijela Horak}{AIG}
\icmlauthor{Simiao Yu}{AIG}
\icmlauthor{Gholamreza Salimi-Khorshidi}{AIG}
\end{icmlauthorlist}

\icmlaffiliation{AIG}{Investments AI, AIG, London, United Kingdom}

\icmlcorrespondingauthor{Danijela Horak}{danijela.horak@aig.com}

\vskip 0.3in
]



\printAffiliationsAndNotice{}  


\begin{abstract}

Automatic evaluation of the goodness of Generative Adversarial Networks (GANs) has been a challenge for the field of machine learning. In this work, we propose a distance complementary to existing measures: Topology Distance (TD), the main idea behind which is to compare the geometric and topological features of the latent manifold of real data with those of generated data. More specifically, we build Vietoris-Rips complex on image features, and define TD based on the differences in persistent-homology groups of the two manifolds. We compare TD with the most commonly-used and relevant measures in the field, including Inception Score (IS), Fr\'echet Inception Distance (FID), Kernel Inception Distance (KID) and Geometry Score (GS), in a range of experiments on various datasets. We demonstrate the unique advantage and superiority of our proposed approach over the aforementioned metrics. A combination of our empirical results and the theoretical argument we propose in favour of TD, strongly supports the claim that TD is a powerful candidate metric that researchers can employ when aiming to automatically evaluate the goodness of GANs’ learning.
\end{abstract}


\section{Introduction}
\label{introduction}

Generative Adversarial Networks (GANs)~\cite{goodfellow_generative_2014} are a class of deep generative models that have achieved unprecedented performance in generating high-quality and diverse images~\cite{brock_large_2019}. They have also been successfully applied to a variety of image-generation tasks, e.g. super resolution~\cite{ledig_photo_2017}, image-to-image translation~\cite{zhu_unpaired_2017}, and text-to-image synthesis~\cite{reed_generative_2016}, to name a few. The GAN framework consists of a generator $G$ and a discriminator $D$, where G generates images $\mathrm{X}_{\mathrm{g}}$ that are expected to resemble real images $\mathrm{X}_{\mathrm{r}}$, while D discriminates between $\mathrm{X}_{\mathrm{g}}$ and $\mathrm{X}_{\mathrm{r}}$. G and D are trained by playing a two-player minimax game in a competing manner. Such novel adversarial training process is a key factor in GANs' success: It implicitly defines a learnable objective that is flexibly adaptive to various complicated image-generation tasks, in which it would be difficult or impossible to explicitly define such an objective.

One of the biggest challenges in the field of generative models -- including for GANs -- is the automatic evaluation of the goodness of such models (e.g., whether or not the data generated by such models are similar to the data they were trained on).
Unlike supervised learning, where the goodness of the models can be assessed by comparing their predictions with the actual labels, or in some other deep-learning models where the goodness of the model can be assessed using the likelihood of the validation data under the distribution that the real data comes from, in most state of the art generative models we do not know this distribution explicitly or can not rely on labels for such evaluations.

Given that the data (or their corresponding features) in such situations can be assumed to lie on a manifold embedded in a high dimensional space~\cite{goodfellow_deep_2016}, tools from topology and geometry come as a natural choice when studying differences between two data set.
Hence, we propose topology distance (TD) for the evaluation of GANs; it compares the the topological structures of two manifolds, and calculates a distance between them to evaluate their (dis)similarities. We compare TD with widely-used and relevant metrics, and demonstrate that it is more robust to noise compared to competing distance measures on GAN's, and it is  better suited to distinguish among various shapes that the data might come in. TD is able to evaluate GANs with new insights different from other existing measurements. It can therefore be used either as an alternative to, or in conjunction with other metrics.


\subsection{Related work}
\label{related_work}

There have been multiple metrics proposed to automatically evaluate the performance of GANs. In this paper we focus on the most commonly-used and relevant approaches (as follows); for a more comprehensive review of such measurements, please refer to~\cite{borji_pros_2018}.

\textbf{Inception Score (IS)} The main idea behind IS~\cite{salimans_improved_2016} is that generated images of high quality are expected to meet two requirements: They should contain easily classifiable objects (i.e. the conditional label distribution $p(\mathrm{y}|\boldsymbol{\mathrm{x}})$ with low entropy) and should be diverse (i.e. the marginal distribution $p(\mathrm{y})$ with high entropy). IS measures the average KL divergence between these two distributions:
\begin{equation}
\mathrm{IS} = \mathrm{exp}(\mathbb{E}_{\boldsymbol{\mathrm{x}} \sim p_{\mathrm{g}}^{}}[\mathrm{KL}(p(\mathrm{y}|\boldsymbol{\mathrm{x}}) \mid \mid p(\mathrm{y}))]),
\end{equation}
where $p_{\mathrm{g}}^{}$ is the generative distribution. IS relies on a pretrained Inception model~\cite{szegedy_rethinking_2016} for the classification of the generated images. Therefore, a key limitation of IS is that it is unable to evaluate the image types that are distinct from those that the Inception model was trained on.

\textbf{Fr\'echet Inception Distance (FID) and Kernel Inception Distance  (KID)} Proposed by~\cite{heusel_gans_2017}, FID relies on a pretrained Inception model, which maps each image to a vector representation (or, features). Given two groups of data in this vector space (one from the real and the other from the generated images), FID measures their similarities, assuming that the features are distributed as multivariate Gaussian; the distance will be the Fr\'echet distance (also known as Wasserstein-2 distance) between the two Gaussians:
\begin{equation}
\mathrm{FID}(p_{\mathrm{r}}^{}, p_{\mathrm{g}}^{}) = \left\lVert \mu_{\mathrm{r}}^{} - \mu_{\mathrm{g}}^{} \right\rVert ^2_2 + \mathrm{Tr}(\Sigma_{\mathrm{r}}^{} + \Sigma_{\mathrm{g}}^{} - 2(\Sigma_{\mathrm{r}}^{}\Sigma_{\mathrm{g}}^{})^{\frac{1}{2}})
\end{equation}
where $p_{\mathrm{r}}^{}$ and $p_{\mathrm{g}}^{}$ denote the feature distributions of real and generated data, ($\mu_{\mathrm{r}}^{}$, $\Sigma_{\mathrm{r}}^{}$) and ($\mu_{\mathrm{g}}^{}$, $\Sigma_{\mathrm{g}}^{}$) denote the means and covariances of the corresponding feature distributions, respectively. It has been shown that FID is more robust to noise (of certain types) than IS~\cite{heusel_gans_2017}, but its assumption of features following a multivariate Gaussian distribution might be an oversimplification.

A similar metric to FID is KID~\cite{binkowski_demystifying_2018}, which computes the squared maximum mean discrepancy (MMD) between the features (learned also from a pretrained Inception model) of real and generated images:
\begin{equation}
\begin{split}
\mathrm{KID}(p_{\mathrm{r}}, p_{\mathrm{g}}) = & 
\mathbb{E}_{\boldsymbol{\mathrm{x}}_{\mathrm{r}}^{}, \boldsymbol{\mathrm{x}}_{\mathrm{r}}' \sim p_{\mathrm{r}}^{}}[k(\boldsymbol{\mathrm{x}}_{\mathrm{r}}^{}, \boldsymbol{\mathrm{x}}_{\mathrm{r}}')] \\ + 
& \mathbb{E}_{\boldsymbol{\mathrm{x}}_{\mathrm{g}}^{}, \boldsymbol{\mathrm{x}}_{\mathrm{g}}' \sim p_{\mathrm{g}}^{}}[k(\boldsymbol{\mathrm{x}}_{\mathrm{g}}^{}, \boldsymbol{\mathrm{x}}_{\mathrm{g}}')] \\ - 
& 2\mathbb{E}_{\boldsymbol{\mathrm{x}}_{\mathrm{r}}^{} \sim p_{\mathrm{r}}^{}, \boldsymbol{\mathrm{x}}_{\mathrm{g}}^{} \sim p_{\mathrm{g}}^{} }[k(\boldsymbol{\mathrm{x}}_{\mathrm{r}}^{}, \boldsymbol{\mathrm{x}}_{\mathrm{g}}^{})] 
\end{split}
\end{equation}
where $k$ denotes a polynomial kernel function $k(\boldsymbol{\mathrm{x}}_{}^{}, \boldsymbol{\mathrm{x}}_{}') = (\frac{1}{\mathrm{d}}\boldsymbol{\mathrm{x}}_{}^\intercal \boldsymbol{\mathrm{x}}_{}'+1)^3$ with feature dimension $\mathrm{d}$. Compared with FID, KID does not have any parametric form assumption for feature distribution, and has a unbiased estimator.

Our proposed TD is closely related to FID and KID in that it also measures the distance between latent features of real and generated data. However, the key distinction of TD is that the target distance is computed by considering the geometric and topological properties of those latent features.

\textbf{Geometry Score (GS)} GS~\cite{khrulkov_geometry_2018} Geometry score 
is defined as 
$l_{2}$ distance between means of the relative living-times (RLT) vectors associated with the two sets of images. RLT of a point cloud data (e.g., a group of images in the feature space) is  an infinite vector $(\mathrm{u}_{1},\mathrm{u}_{2},\ldots)$ whose $\mathrm{i}$-th entry is a measure of persistent intervals having $1$-persistent homology group rank equal to $\mathrm{i}$. That is,
$\mathrm{u}_{\mathrm{i}}=\frac{1}{\mathrm{d}_{\mathrm{n}(\mathrm{n}-1)/2}} \sum_{\mathrm{j}=1}^{\mathrm{n}(\mathrm{n}-1)/2} \mathrm{I}_{\mathrm{j}}(\mathrm{d}_{\mathrm{j}+1}-\mathrm{d}_{\mathrm{j}})$, where 
$\mathrm{I}_{\mathrm{j}}$ equals $1$ if the rank of persistent homology group of dimension $1$ in interval $[\mathrm{d}_{\mathrm{j}},\mathrm{d}_{\mathrm{j}+1}]$ is $\mathrm{i}$, and zero otherwise. Persistent homology parameters $\mathrm{d}_{\mathrm{i}}$, $\mathrm{i}\in [0\dots \mathrm{n}(\mathrm{n}-1)/2]$ are sorted distances in the observed point cloud data. 

Geometry score exploits similar idea to the topological distance, with the difference being in  the underlying point cloud data used, dimensionality of the homology group and distance measure between the persistent diagrams.  We claim that our method better aligns with the existing theory in the area of computational algebraic topology and has superior experimental results.


\section{Main idea}
\label{main_idea}
According to the manifold hypothesis~\cite{goodfellow_deep_2016}, real world high dimensional data (and their features) lie on a low dimensional manifold embedded in a high dimensional space. The main idea of this paper is to compare the latent  manifold of the real data with that of the generated data, based entirely on the topological properties of the data samples from these two manifolds. Let $\mathbb{M}_{\mathrm{r}}$ and $\mathbb{M}_{\mathrm{g}}$ be the latent  manifolds of the real and generated data, respectively. We aim to compare these two manifolds using the finite samples of points $\mathrm{V}_{\mathrm{r}}$ from $\mathbb{M}_{\mathrm{r}}$ and $\mathrm{V}_{\mathrm{g}}$ from $\mathbb{M}_{\mathrm{g}}$.

Most mainstream methods compare samples $\mathrm{V}_{\mathrm{r}}$  and $\mathrm{V}_{\mathrm{g}}$ using the lower order moments (e.g.~\cite{heusel_gans_2017}) -- similar to the way we compare two functions using their Taylor expansion, for instance.
However, this would only be valid  if the underlying manifold is an Euclidean space (zero curvature), as all moments of the samples are calculated using Euclidean distance. For a Riemannian manifold with a nonzero curvature, this type of approach, at least in theory, would not work, and using geodesic instead of Euclidean distance would agree more with the hypothesis. 

Here we propose the comparison of the two manifolds on the basis of their topology and/or geometry. The ideal way to compare two manifolds would be to infer if they are geometrically equivalent, i.e. isometric. This, unfortunately, is not attainable. However, we could compare two manifolds by the means of eigenvalues of the Laplace-Beltrami operator\footnote{Only for Riemannian manifolds} on them.

The Laplace-Beltrami spectrum can be regarded as the set of squared frequencies that are associated to the modes of eigenvalues of an oscillating membrane defined on the manifold. The spectrum then, is an infinite sequence of eigenvalues, and satisfies some nice stability properties, whereby a small perturbation in the metric of the underlying Riemannian manifold results in a small perturbation of the spectrum~\cite{donnelly_spectral_2010, birman_on_1963}. Furthermore, the Laplace-Beltrami spectrum is widely considered as a ``fingerprint'' of a manifold. In 1966, in the famous paper ``Can one hear the shape of a drum?''~\cite{kac_can_1966}, M. Kac has asked  a question whether the eigenvalues of Laplace Beltrami operator alone are sufficient to uniquely (up to an isometry) identify a manifold. The answer is unfortunately not, but the isospectral manifolds are rare and when they exist, they share multiple topological and geometric features.

Furthermore, it is possible to translate this methodology to a discrete setting, such that the spectrum calculated on the discrete set relates closely to the spectrum on the manifold itself. 

\textbf{Theorem 1}~\cite{mantuano_discretization_2005}
\textit{ 
Given a discretisation $\mathrm{G}= \mathrm{G}(\mathrm{V}, \mathrm{\epsilon})$ of a compact Riemannian manifold $\mathbb{M}$ which has non-negative sectional curvature $\kappa$, and non negative injectivity radius, and for which  $Ricci(\mathbb{M},\mathrm{g})\geq -(\mathrm{n}-1)\mathrm{\kappa} \mathrm{g}$, where $\mathrm{n}$ is dimension of a manifold, $\mathrm{g}$ is a Riemannian metric, then it is possible to associate the eigenvalues of Laplace operator on a graph $\mathrm{G}$, with the ones of the  Laplace Beltrami operator on $\mathbb{M}(\mathrm{c}_{1}\lambda_{\mathrm{k}}(\mathrm{G})\leq\lambda_{\mathrm{k}}(\mathbb{M})\leq\mathrm{c}_{2}\mathrm{\lambda}_{\mathrm{k}}(\mathrm{G}))$, for all $k<\lvert V \lvert$.}

The discretisation  $\mathrm{G}(\mathrm{V}, \mathrm{\epsilon})$ of a manifold $\mathbb{M}$, is a set of points in $\mathbb{M}$ whose distance is at least $\mathrm{\epsilon}$  and the union of the balls centred in the points of $\mathrm{V}$ with radius $\mathrm{\epsilon}$ which forms an open cover of $\mathbb{M}$, denoted by $\mathcal{U}$. A version of this theorem also holds for eigenvalues of higher dimensional version of Laplace Beltrami operator, called  Laplace–de Rham operator which reflects high dimensional topological and geometric properties of a manifold \cite{mantuano_discretization_2008}. 
This effectively means that for a sufficiently good sample $\mathrm{V}$ from $\mathbb{M}$, we can claim that calculating the eigenvalues on $\mathrm{V}$ would effectively be as calculating them on $\mathbb{M}$ (see~\cite{dey_convergence_2010} for more results).

In our case the manifold $\mathbb{M}$ is unknown, and all we know is a sample of points from it: $\mathrm{V}$. In order to calculate the Laplace-Beltrami spectrum, we need to have a graph structure on $\mathrm{V}$, which comes through \v{C}ech complex on its cover $\mathcal{U}$. To obtain the \v{C}ech complex, one needs radii of the balls in the cover, i.e., $\mathrm{\epsilon}$. This in itself poses a problem, because it is difficult to determine the right value of $\mathrm{\epsilon}$. There is little hope in recovering spectral properties of $\mathbb{M}$ from the point sample $\mathrm{V}$, because we are unable to determine the right value of  $\mathrm{\epsilon}$.

A similar theorem to Theorem 1 applies to homology type of a manifold and its sample.

\textbf{Theorem 2} 
\textit{
Given a  Riemannian manifold $\mathbb{M}$, and a sample of points from it, $\mathrm{V}$, which  is sufficiently dense, then a  Vietoris–Rips complex of $\mathrm{V}$ at scale $\mathrm{\epsilon}$ is homologically equivalent to $\mathbb{M}$.
}

This theorem is a direct consequence of the famous nerve theorem \cite{alexandroff_uber_1928}, but can also be seen as a consequence of Theorem 1, due to a fact that the multiplicity of eigenvalue zero on discretised space is exactly the rank of a homology group of dimension zero on the same spacee. 

In practice, as before, one does not know how to choose scale for $\mathrm{\epsilon}$, but unlike before, in this setting we have available a tool that can, and is specifically designed to, deal with the uncertainty of scale: persistent homology. We chose to utilise persistent homology to extract information about the geometry and topology of $\mathbb{M}$, because persistent homology, measured on the sample $\mathrm{V}$, is a reliable shape quantifier of $\mathbb{M}$.


\section{Preliminaries}
\label{preliminaries}

Intuitively speaking, topological space is any space on which the notion on neighbourhood can be defined. Hence, all metric spaces (and consequently all examples considered in this work) are topological spaces; the opposite is not true (i.e., not all topological spaces can be endowed with a metric).

It is very difficult to directly assess whether two topological spaces are equivalent (homeomorphic); instead topologists use proxies to measure their similarity.
One of these proxies are homology groups, denoted by $\mathrm{H}_{\mathrm{k}}, \mathrm{k}\in \mathbb{N}_{0}$, which loosely speaking encode the information on different types of loops (of different dimensions) that can be observed in the topological space. And here, the following implication holds: If two topological spaces are equivalent, then their homology groups are isomorphic, but the opposite is not true.

In this work we will only be concerned with a special class of topological spaces called simplicial complexes. Simplicial complex, commonly denoted by $\mathrm{K}$,  is a  topological space consisting of vertices in a set $\mathrm{V}$ and a set of faces chosen from the partitive set of $\mathrm{V}$, $\mathcal{P}(\mathrm{V})$, with the requirement that if $\mathrm{W}\in \mathrm{K}$, then all the subsets of $\mathrm{W}$ are also in $\mathrm{K}$. One way to visualise the simplicial complex is to consider vertices as points in $\mathbb{R}^{\mathrm{n}}$ and $\mathrm{m}$-dimensional faces as convex hulls of $\mathrm{m}+1$ vertices, i.e. edges, triangles, tetrahedra, etc.

Homology groups are algebraic constructions defined by
\begin{equation}
    \mathrm{H}_{\mathrm{k}}(\mathrm{K},\mathbb{R})=\mathrm{Z}_{\mathrm{k}}(\mathrm{K},\mathbb{R})/\mathrm{B}_{\mathrm{k}}(\mathrm{K},\mathbb{R}),
\end{equation}
where $\mathrm{k}$ is a non-negative integer, $\mathrm{Z}_{\mathrm{k}}(\mathrm{K},\mathbb{R})$ is a vector space of $\mathrm{k}$-dimensional cycles and $\mathrm{B}_{\mathrm{k}}(\mathrm{K},\mathbb{R})$ a $\mathrm{k}$-dimensional boundaries, obtained as per-images and images of the boundary mapping on a chain complex,  for more detailed account see~\cite{hatcher_algebraic_2009}. Typically a rank of a homology group of dimension zero would be the number of connected components of $\mathrm{K}$, rank of $\mathrm{H}_{1}$ would be the number of one dimensional holes in $\mathrm{K}$, rank of $\mathrm{H}_{2}$ would be the number of cavities, and so on.

A persistent homology, loosely speaking, is a homology of a topological space measured at different resolutions. More precisely,  we study a nested sequence of topological spaces (i.e., filtration $\mathcal{K}: \mathrm{K}_{0}\subset \mathrm{K}_{1}\subset \ldots\subset \mathrm{K}_{\mathrm{N}}$) and measure (calculate) homology at every step. As an example let's observe the sequence of Vietoris-Rips complexes on a point set $\mathrm{V}$: A Vietoris-Rips complex on a vertex set $\mathrm{V}$ and a diameter $\mathrm{\epsilon}$ is a simplicial complex,in which   $\mathrm{v}_{0}, \ldots, \mathrm{v}_{k}$ is a simplex iff $d(\mathrm{v}_{\mathrm{i}},\mathrm{v}_{\mathrm{j}})<\mathrm{\epsilon}$ for every $\mathrm{i},\mathrm{j}\leq \mathrm{k}$.

\begin{figure}[t]
  \includegraphics[width=\linewidth]{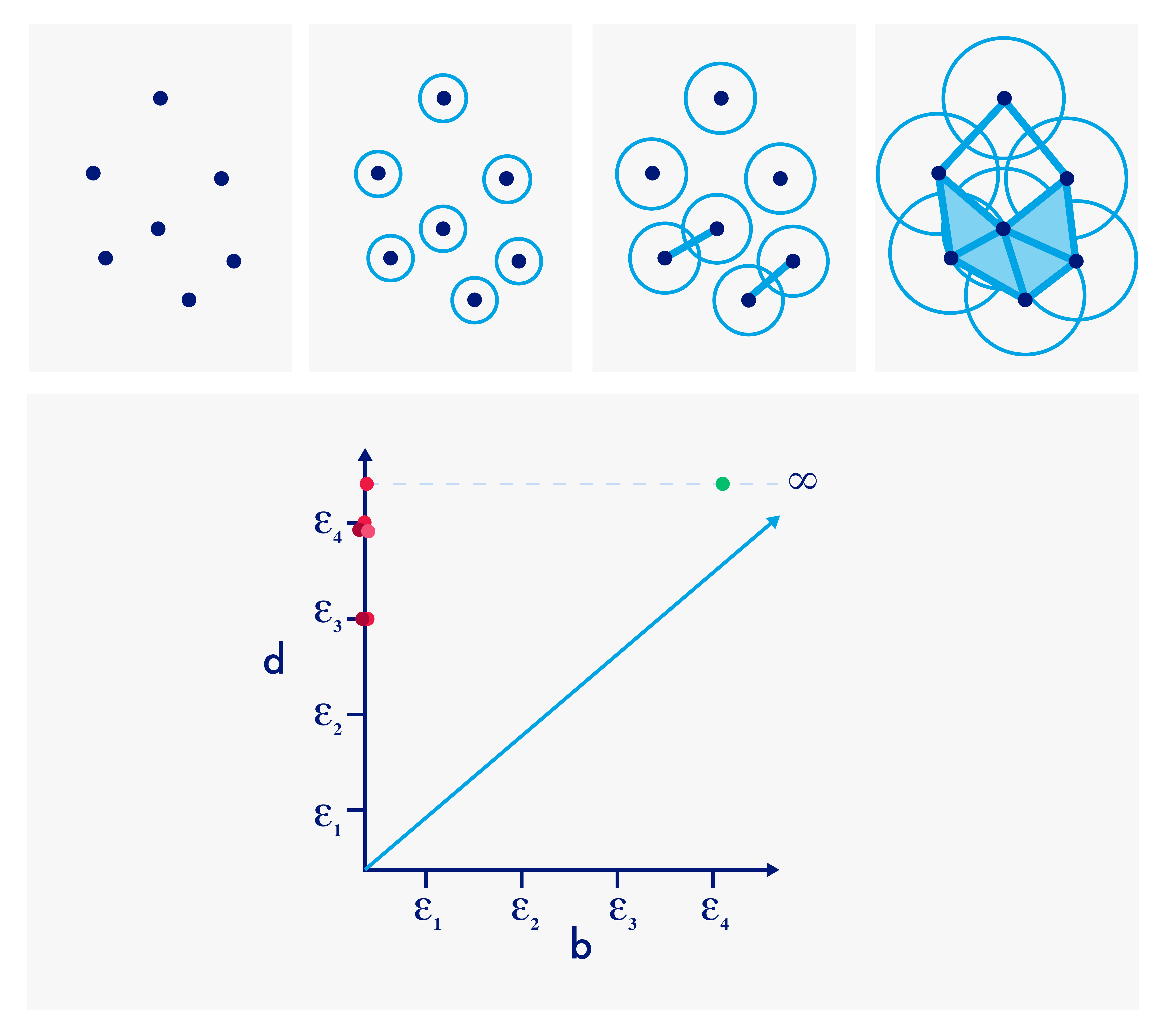}
  \caption{ (top) The $4$ step filtration of Vitoris Rips complex on the set of $7$ points with increasing radius $0=\varepsilon_{1}\leq \ldots\leq \varepsilon_{4} $.   (bottom) The persistent diagram corresponding to the filtration in the figure on top: $b$-axis denotes "birth" or appearance of persistent homology group, and $d$-axis denotes "death" or disappearance. Red points represent persistent homology groups of dimension $0$, and the green ones of dimension $1$.}
  \label{fig:VRips}
\end{figure}

An example of a filtration would be 
$\mathcal{VR}: \mathrm{VR}(\mathrm{V},\epsilon_{0})\subseteq \mathrm{VR}(\mathrm{V},\epsilon_{1})\subseteq \ldots\subseteq \mathrm{VR}(\mathrm{V},\epsilon_{\mathrm{n}})$, where 
$0=\epsilon_{0}\leq \epsilon_{1}\leq \ldots\leq \epsilon_{\mathrm{n}} $ (see Figure~\ref{fig:VRips} (top)).
In other words, persistent homology quantifies a change of topological invariants in  $\mathcal{\mathrm{VR}}$ with a change of parameter $\mathrm{\epsilon}$.

Formally, 
\begin{equation}
\mathrm{H}_{\mathrm{k}}^{\mathrm{i},\mathrm{j}}(\mathcal{VR})=\mathrm{Z}_{\mathrm{k}}^{\mathrm{i}}/\mathrm{B}_{\mathrm{k}}^{\mathrm{j}}\cap \mathrm{Z}_{\mathrm{k}}^{\mathrm{i}},
\end{equation}
where the $\mathrm{j}$-th persistent homology group of dimension $\mathrm{k}$ of the $\mathrm{i}$-th filtration complex $\mathrm{VR}(\mathrm{V}, \epsilon_{\mathrm{i}})$ is denoted by $\mathrm{H}_{\mathrm{k}}^{\mathrm{i},\mathrm{j}}(\mathcal{VR})$. Intuitively, the persistent homology group records the ``cycles'' at the filtration step $\mathrm{i}$, which have not become ``boundaries'' (i.e. which have not effectively disappeared ) at filtration step $\mathrm{j}$. For detailed account see~\cite{edelsbrunner_persistent_2008}.

The main insight when it comes to persistent homology is that the evolution of topological invariants  over increase in parameter $\mathrm{\epsilon}$, can be encoded compactly in the form of  a persistent diagram  $\mathcal{PD}$ and a barcode.
\begin{equation}
\mathcal{PD}_{\mathrm{k}}( \mathcal{VR})=\{(\mathrm{b}_{\mathrm{i}},\mathrm{d}_{\mathrm{i}})\mid {\mathrm{i}}\in \mathbb{N}, \mathrm{b}_{\mathrm{i}},\mathrm{d}_{\mathrm{i}}\in \{\mathrm{\epsilon}_{0},\ldots,\mathrm{\epsilon}_{\mathrm{n}}\}\},
\end{equation}
where $\mathrm{b}_{\mathrm{i}}$ in the pair $((\mathrm{b}_{\mathrm{i}},\mathrm{d}_{\mathrm{i}}))$ records the appearance (or birth) of a $\mathrm{k}$-dimensional homology group and $\mathrm{d}_{\mathrm{i}}$ records its disappearance (also referred to as "death"). In the event that homology group persists, i.e. it does not disappear during the end of filtration, we set $\mathrm{d}_{\mathrm{i}}=\infty$. 
This set of points is represented in the upper triangle of the first quadrant of the $\mathbb{R}^{2}$ (see Figure~\ref{fig:VRips} (bottom)).
Another, representation is a barcode where each bar is mapped to a point $((\mathrm{b}_{\mathrm{i}},\mathrm{d}_{\mathrm{i}}))$  with the starting point $\mathrm{b}_{\mathrm{i}}$ and ending point $\mathrm{d}_{\mathrm{i}}$.

There is a natural measure of distance defined on persistent diagrams, 
$\infty$-Wasserstein distance, also known in the community as the bottleneck distance, which has desirable stability properties with respect to small perturbations~\cite{chazal_the_2016}, but is sensitive to outliers and mostly unsuitable for use in practice. 
On the other hand, $\mathrm{p}$-Wasserstein distance
\begin{equation}
\begin{split}
\mathrm{W}_{\mathrm{q}}(\mathrm{L}_{\mathrm{p}})(\mathcal{PD}^{0}_{k}, \mathcal{PD}^{1}) &= \\
\Big[\inf_{\mathrm{\eta}:\mathcal{PD}^{0}_{k}\rightarrow \mathcal{PD}^{1}_{k}}&\sum_{\mathrm{u}\in \mathcal{PD}^{0}_{k}}\parallel \mathrm{u}-\mathrm{\eta}(\mathrm{u})\parallel_{\mathrm{q}}\Big]^{1/\mathrm{p}},
\end{split}
\end{equation}
where $1\leq \mathrm{p}, \mathrm{q} \leq \infty$, and   $\mathrm{\eta}$ ranges over all bijections between sets of persistent intervals in diagrams $\mathcal{PD}_{0}$ and $\mathcal{PD}_{1}$, shows more potential as presented in~\cite{chazal_robust_2018}, but is computationally demanding.

In practice, much of the applications of persistent homology have used neither of the two distances, but have  relied on ad-hoc distances between persistent diagrams, which do not have a strong backing in theory (e.g.~\cite{bendich_persistent_2016, khrulkov_geometry_2018}.

To conclude, endowed with any distance measure described above, the space of persistent diagrams is a metric space.


\section{Method}
\label{method}

The method we propose for evaluation of the performance of  generative models rests on measuring the differences between the set of images generated by GANs and set of original images.
We measure the distance on the point cloud data in feature space.
Let $\mathrm{F}_{\mathrm{r}}$ be the set of features of the real, and $\mathrm{F}_{\mathrm{g}}$ the set of features of the generated images represented in feature space: $\mathbb{R}^{\mathrm{m}}$.

Seen as the point cloud data in $\mathbb{R}^{\mathrm{m}}$, one can calculate the distances between the points in $\mathrm{F}_{\mathrm{r}}$. 
It is worth noting here again, that even though we calculate all  the distances using Euclidean metric, the algorithm will effectively use only  "small" distances, and this is in agreement with potential non-zero curvature of the manifold(refer to Theorem 1 and Theorem 2 for full statement of this fact).

Assume that there are $\mathrm{n}$ data points in $\mathrm{F}_{\mathrm{r}}$ and $\mathrm{F}_{\mathrm{g}}$, and  let $0=\mathrm{d}^{\mathrm{r}}_{0}\leq \mathrm{d}^{\mathrm{r}}_{1}\leq \ldots \leq \mathrm{d}^{\mathrm{r}}_{\mathrm{n}(\mathrm{n}-1)/2}$  and $0=\mathrm{d}^{\mathrm{g}}_{0}\leq \mathrm{d}^{\mathrm{g}}_{1}\leq \ldots \leq \mathrm{d}^{\mathrm{g}}_{\mathrm{n}(\mathrm{n}-1)/2}$ be an array of sorted distances among vectors in $\mathrm{F}_{\mathrm{r}}$, $\mathrm{F}_{\mathrm{g}}$, respectively.
Then we observe the following filtration:
$\mathcal{VR} (\mathrm{F}_{\mathrm{r}}): \mathrm{VR}(\mathrm{F}_{\mathrm{r}},\mathrm{d}^{\mathrm{r}}_{0})\subseteq \ldots\subseteq \mathrm{VR}(\mathrm{F}_{\mathrm{r}},\mathrm{d}^{\mathrm{r}}_{\mathrm{k}})$ and $\mathcal{VR}(\mathrm{F}_{\mathrm{g}}): \mathrm{VR}(\mathrm{F}_{\mathrm{g}},\mathrm{d}^{\mathrm{g}}_{0})\subseteq \ldots\subseteq \mathrm{VR}(\mathrm{F}_{\mathrm{g}},\mathrm{d}^{\mathrm{g}}_{\mathrm{l}})$, where the distance $\mathrm{d}^{\mathrm{r}}_{\mathrm{k}}$ is the minimal distance $\mathrm{d}$ for which corresponding Vietoris Rips complex becomes fully connected. Same is true for  $\mathrm{d}^{\mathrm{r}}_{\mathrm{l}}$.
The $0$ dimensional persistent homology groups are  calculated on the aforementioned filtrations.
One consequence of studying only $0$th dimensional persistent homology group, is that the rank of the persistent homology group at time $0$ will be exactly $\mathrm{n}$, and persistent diagram will consist of $\mathrm{n}$ pairs $(\mathrm{b}_{\mathrm{i}},\mathrm{d}_{\mathrm{i}})$, where 
$\mathrm{b}_{\mathrm{i}}$ denotes the point in filtration where the observed homology group has appeared for the first time (In our case $\mathrm{b}_{\mathrm{i}}=0$, for every $\mathrm{i}$, due to the choice of filtration), and $\mathrm{d}_{\mathrm{i}}$ denotes a point in filtration where the observed homology group(connected component) has merged with another one, or is equal to $\infty$ otherwise.
This observation holds for both $\mathrm{F}_{\mathrm{r}}$ and $\mathrm{F}_{\mathrm{g}}$.

\begin{algorithm}[t]
	\caption{This algorithm is to compute the longevity vector $l$ for a set of images. $l$ is of length $\mathrm{n}$, which represents living times of all $\mathrm{n}$ homology classes throughout filtration (see Section~\ref{method} for more details).}
	\label{alg1}
	\begin{algorithmic}
		\STATE {\bfseries Require:} $f^{*}_{\theta}$: a pretrained feature extractor with parameters $\theta$.
        \STATE {\bfseries Require:} $RC(\mathrm{p})$: a function for computing Vietoris-Rips Filtration over the given input points $\mathrm{p}$.
        \STATE {\bfseries Require:} $PD(\mathrm{c})$: a function for computing persistent homology  in dimension 0 of filtration $\mathrm{c}$. 
        \STATE {\bfseries Input:} $\mathrm{X} \in \mathbb{R}^{\mathrm{N} \times \mathrm{H} \times \mathrm{W} \times \mathrm{C}}$: a set of images of size $\mathrm{H} \times \mathrm{W}$ with $\mathrm{C}$ channels.
        \newline
		\STATE {\bfseries Compute:} $\mathrm{F} \leftarrow f^{*}_{\theta}(\mathrm{X})$ \\
		\STATE {\bfseries Compute:} $\mathrm{C} \leftarrow RC(\mathrm{F})$ \\		
		\STATE {\bfseries Compute:} $l \leftarrow PD(\mathrm{C})$ \\	
	\end{algorithmic}
\end{algorithm}

\begin{algorithm}[t]
	\caption{This algorithm is to compute Topology Distance (TD) between real and generated images.}
	\label{alg2}
	\begin{algorithmic}
        \STATE {\bfseries Input:} $\mathrm{X}_{\mathrm{r}} \in \mathbb{R}^{\mathrm{N} \times \mathrm{H} \times \mathrm{W} \times \mathrm{C}}$: a set of real images of size $\mathrm{H} \times \mathrm{W}$ with $\mathrm{C}$ channels.
        \STATE {\bfseries Input:} $\mathrm{X}_{\mathrm{g}} \in \mathbb{R}^{\mathrm{N} \times \mathrm{H} \times \mathrm{W} \times \mathrm{C}}$: a set of generated images of size $\mathrm{H} \times \mathrm{W}$ with $\mathrm{C}$ channels.
        \newline
		\STATE {\bfseries Compute:} $l_{\mathrm{r}}$ with $\mathrm{X}_{\mathrm{r}}$ using Algorithm~\ref{alg1}.
		\STATE {\bfseries Compute:} $l_{\mathrm{g}}$ with $\mathrm{X}_{\mathrm{g}}$ using Algorithm~\ref{alg1}.		
		\STATE {\bfseries Compute:} $TD(\mathrm{X}_{\mathrm{r}}, \mathrm{X}_{\mathrm{g}}) \leftarrow \parallel l_{\mathrm{r}}-l_{\mathrm{g}}\parallel_2$					
	\end{algorithmic}
\end{algorithm}

As mentioned in Section~\ref{preliminaries}, a commonly used distance between persistence diagrams in the field of topological data analysis is bottleneck distance as it has shown desirable properties of stability.
However, we have found that this distance is too sensitive to outliers and its practical applications are limited. Hence, we've chosen to define the distance between the persistence diagrams differently. In fact we've used the inherent properties of our filtration method.
We assign a $\mathrm{n}$-dimensional vector $l(\mathrm{F}_{\mathrm{r}})= (\mathrm{d}_{0}-\mathrm{b}_{0},\ldots, \mathrm{d}_{\mathrm{n}}-\mathrm{b}_{\mathrm{n}})$, called the \textbf{longevity vector} to the persistent diagram which represents the sorted living times of each homology group for point set $\mathrm{F}_{\mathrm{r}}$. Same for $\mathrm{F}_{\mathrm{g}}$.

Then, we define the Topology Distance (TD) between two persistent diagrams, and consequently between two corpuses of images to be $\mathrm{l}_{2}$ distance between their longevity vectors, i.e. 
$\mathrm{TD}(\mathrm{F}_{\mathrm{r}}, \mathrm{F}_{\mathrm{g}}) =  \parallel l(\mathrm{F}_{\mathrm{r}})-l(\mathrm{F}_{\mathrm{g}})\parallel_{2}$, where $l(\mathrm{F}_{\mathrm{r}})$ and $l(\mathrm{F}_{\mathrm{g}})$ are the longevity vectors of persistent diagrams of filtrations of set of original and generated image features, respectively. 

Furthermore, as some persistent pairs may contain $\infty$ we will assume that the difference between two infinite coordinates in $0$, and the difference between $\infty$ and non-infinite coordinate in our algorithm is a some fixed value larger than the maximum finite longevity. 

Our method to translate persistent diagrams in feature vectors is indeed ad-hoc, but the justification comes through the fact that  neither bottleneck distance nor Wasserstein distance showed satisfactory experimental results. The TD algorithm is summarised in Algorithm~\ref{alg1} and Algorithm~\ref{alg2}. 


\section{Experiments}
\label{experiments}


\subsection{Datasets and experimental setup}

We compared our proposed TD (lower is better) with IS (higher is better), FID (lower is better), KID (lower is better) and GS (lower is better) as introduced in Section~\ref{related_work}. In addition to some simulated data, which we will introduce in the next Section, our experiments were carried out on the following four datasets: Fashion-MNIST~\cite{xiao_fashionmnist_2017}, CIFAR10~\cite{krizhevsky_learning_2009}, corrupted CIFAR100 (CIFAR100-C)~\cite{hendrycks_benchmarking_2019} and CelebA~\cite{liu_deep_2015}. Wherever features were required for computing the metric, we used a ResNet18 model~\cite{he_deep_2016} trained from scratch for Fashion-MNIST images, and the Inception model~\cite{szegedy_rethinking_2016} pretrained on ImageNet~\cite{deng_imagenet_2009} for all other datasets.

We implemented our algorithm using Python version of GUDHI\footnote[1]{http://gudhi.gforge.inria.fr/} for topology-related computation and PyTorch~\cite{paszke_pytorch_2019} for building and training neural network models.

\begin{figure}[tbh]
  \includegraphics[width=\linewidth]{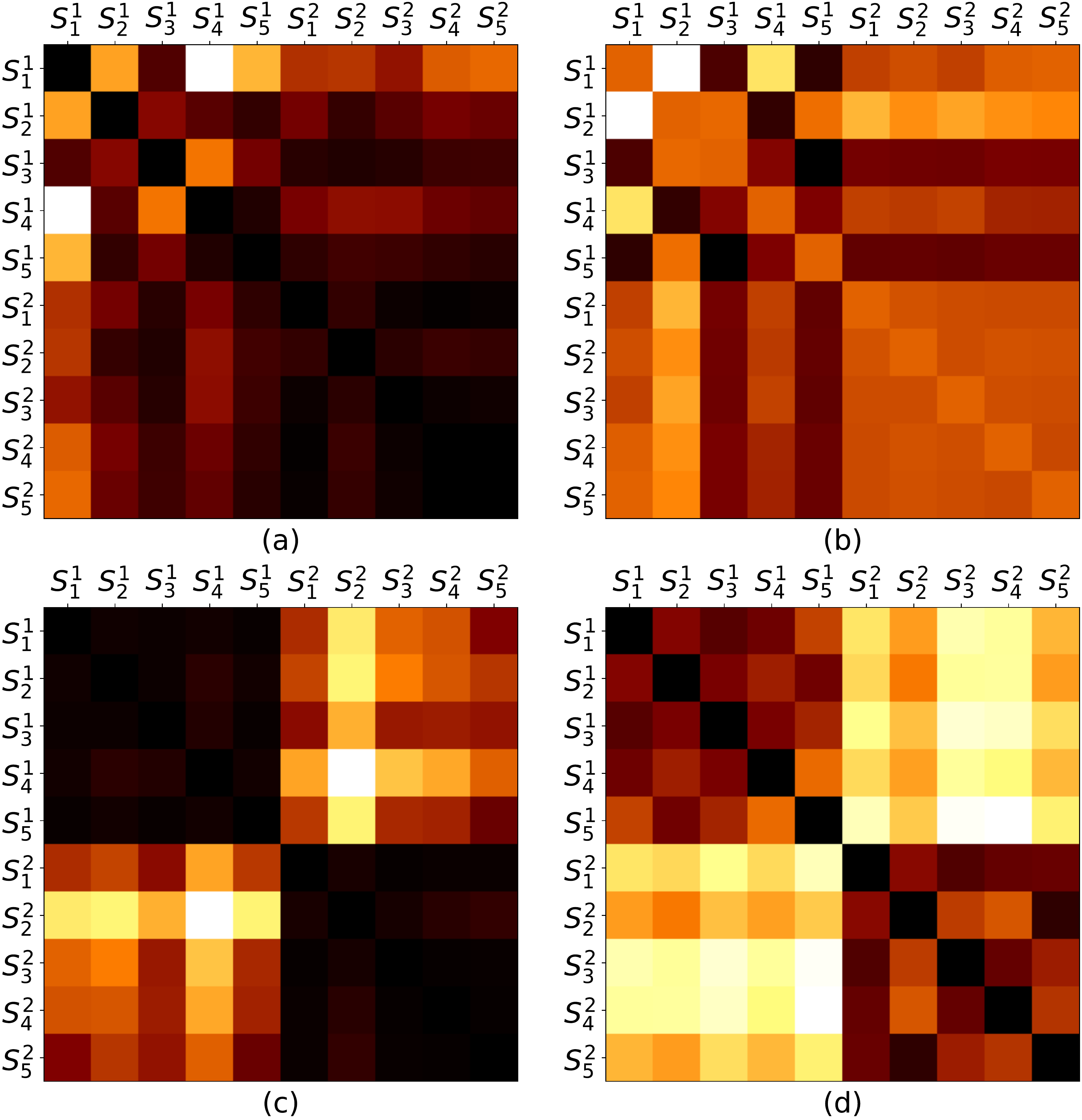}
  \caption{Heat maps of distance matrices between 5 sample sets (each of which has 600 samples randomly sampled from a single Gaussian distribution, denoted by $s_{i}^{1}$), and another 5 sample sets (each containing 600 samples from a mixture of two Gaussian distributions, denoted by $s_{i}^{2}$). All $s_{i}^{j}$s have the same first and second moments. (a)  FID. (b) KID. (c) GS. (d) TD.}
  \label{fig:synthetic_gaussians}
\end{figure}


\subsection{Comparison with FID and KID}

The idea of basing the distance measure entirely on the first two moments (e.g., {\em a la} FID) can be an oversimplification of the underlying distributions at times, as describing certain distributions require the use of higher order statistics (e.g., third or fourth moments).
Furthermore, if two distributions have identical moments of all orders, it is still possible for them to be different distributions~\cite{romano_counterexamples_1986}. This leads to a conclusion that any distance metric based entirely on moments cannot successfully distinguish between all probability distributions. 

\begin{figure}[tbh]
  \includegraphics[width=\linewidth]{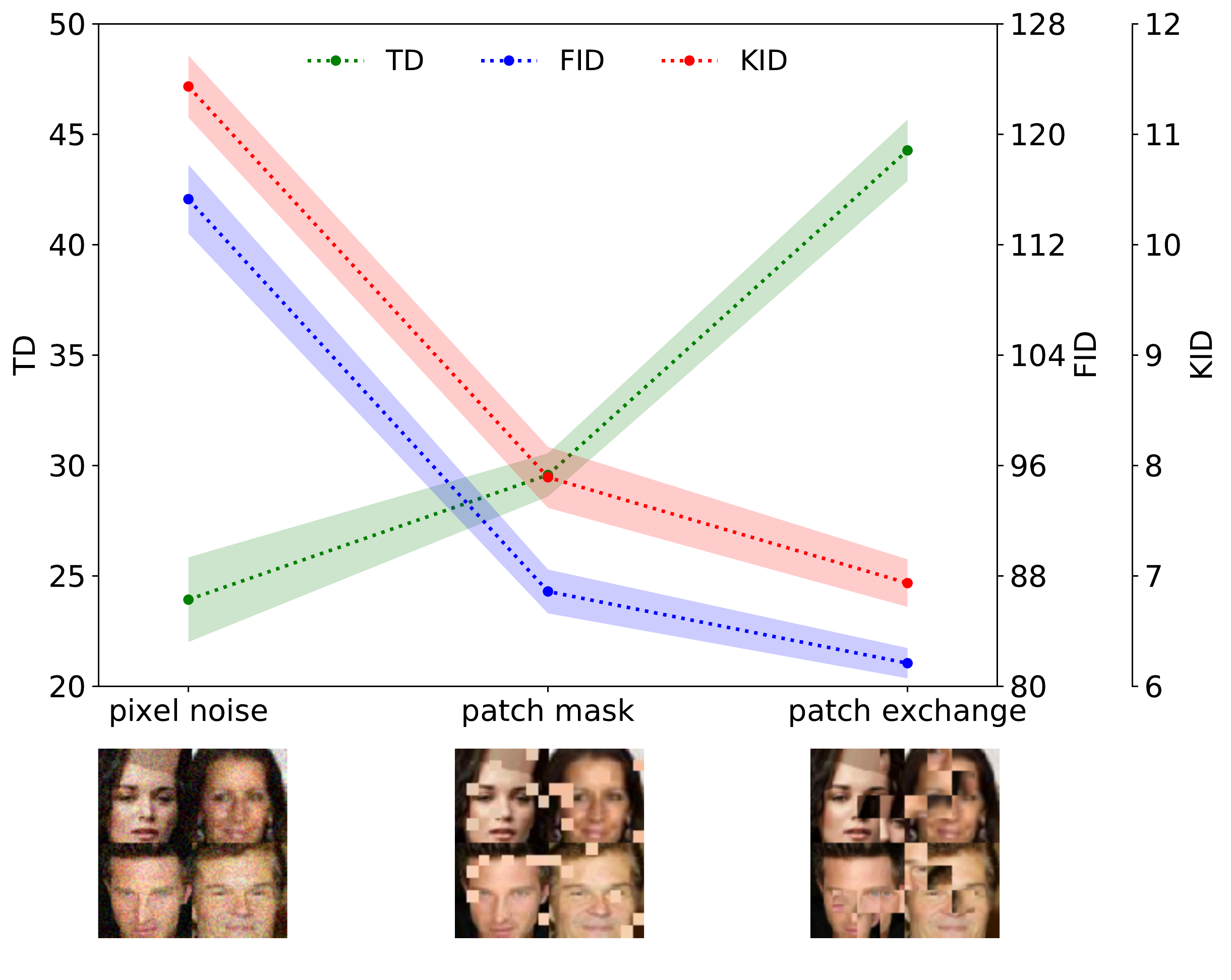}
  \caption{Comparison of FID, KID ($\times100$) and TD on manipulated images (with pixel noise, patch mask and patch exchange). Results are averaged over 10 groups, each consisting of 500 real images (randomly sampled from the CelebA dataset) and the corresponding manipulated counterparts.}
  \label{fig:fid_vs_td_celeba}
\end{figure}

In order to assess how such theoretical considerations will affect FID score's performance, we first compared TD and FID on a synthetic dataset. As shown in Figure~\ref{fig:synthetic_gaussians} we aim to calculate the distance between a single Gaussian distribution and a mixture of two Gaussian distributions (the mixture has the same mean and variance as the single Gaussian).
Given the identical first and second moments of the two point clouds in this case, as expected, FID cannot discriminate between the two, whereas the difference is very obvious when using TD. KID has similar limitations as FID, as demonstrated in Figure~\ref{fig:synthetic_gaussians}.


\begin{figure*}[tbh]
  \includegraphics[width=\linewidth]{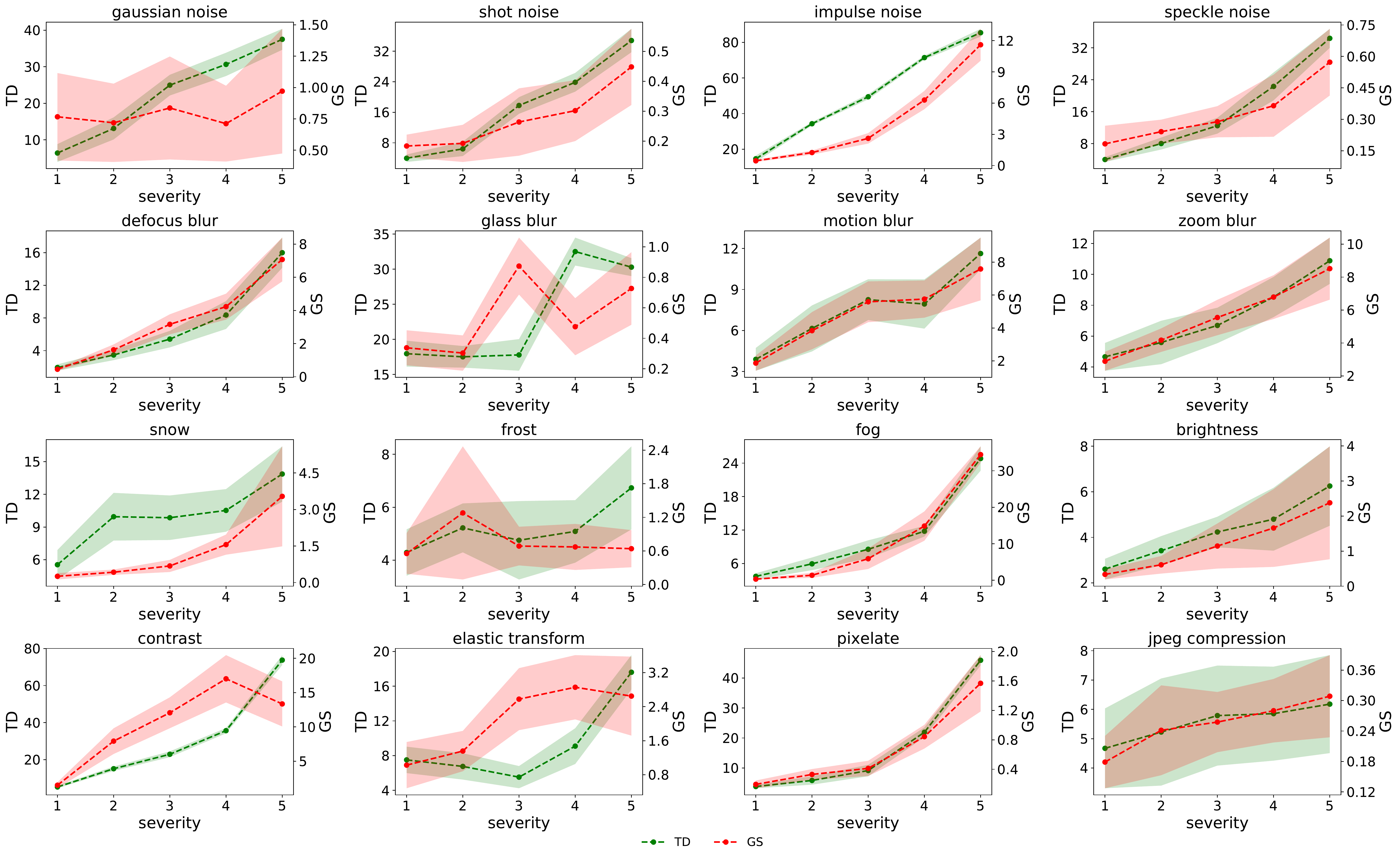}
  \caption{Comparison of perturbation consistency between TD and GS ($\times$1000) on CIFAR100-C dataset. Each row represents one group of perturbation (i.e. noise, blur, weather and digital), and there are four specific perturbation types in each group. Results are averaged over 10 groups, each of which consists of 500 real images (randomly sampled from the CIFAR100-C dataset) and the corresponding images with perturbation.}
  \label{fig:noise_consistency}
\end{figure*}

Next, we compared TD with FID and KID on real images, randomly sampled from CelebA dataset; the goal was to compare the actual images with their manipulated counterparts. More specifically, we performed three types of manipulations designed by~\cite{liu_improved_2018}, which resulted in three new image datasets; we then computed the distance between each one of these manipulated image datasets and the original image dataset, using TD, FID and KID.

The three image manipulations include: 1) pixel noise (i.e., adding a random noise to each pixel, where the noise is uniformly sampled from the following interval: $1\pm0.13$ times the maximum intensity of the image), 2) patch mask (7 out of 64 evenly-divided regions of each image were masked by a random pixel from the image), and 3) patch exchange (2 out of 16 evenly-divided regions of each image were randomly exchanged with each other, performed twice). Some example images after manipulation are shown in Figure~\ref{fig:fid_vs_td_celeba}. It is clear that the image quality increasingly worsens as we go from pixel noise, to patch mask and patch exchange; we expect to see this trend in the metrics. 

However, as presented in Figure~\ref{fig:fid_vs_td_celeba}, FID and KID show a decreasing trend (indicating increasingly better quality) over pixel noise, patch mask and patch exchange, which is apparently opposite the human judgements. In other words, they fail to capture the worsening of image quality, as expected from qualitative assessments presented in~\cite{liu_improved_2018}; unlike TD, which captures the change in the quality of the manipulated images very well. Since all three metrics are based on the features extracted by the same Inception model, this experiment demonstrates that the superiority of TD over FID and KID is due to its effective assessment of the topological properties of the point clouds (rather than their lower-order statistics).


\subsection{Comparison with GS}
So far, we have attempted to demonstrate the effectiveness of topology in assessing the (dis)similarities of two point clouds. On the other hand, as noted earlier, both topology distance and geometry score exploit the idea of using topology to quantify dissimilarities between the latent manifolds of data. There are, however, two major differences between TD and GS. The first one is in the core method and the way topology is used to construct the distance (for more details see Section~\ref{method}). The second one is that TD measures distances between point could data in the feature space, whereas geometry score is defined on raw pixels. 

Figure~\ref{fig:synthetic_gaussians}c shows the heatmap of the distance matrix calculated between a single Gaussian distribution and a mixture of two Gaussian distributions using GS. It is clear that TD better discriminates between the samples from the two aforementioned distributions (see Figure~\ref{fig:synthetic_gaussians}d).

We then performed perturbation consistency comparison between TD and GS using the CIFAR100-C dataset, in which 16 different types of perturbations (grouped in four, namely noise, blur, weather and digital) are applied to the original CIFAR100 images; for each type of perturbation there are five levels of severity. 5,000 images were randomly sampled from the real dataset, and split into 10 groups (each with 500 images); for each group, scores are calculated comparing the perturbed images and the original. For every perturbation, as severity increases, the average score (across 10 groups) should increase monotonically with it.

As can be seen in Figure~\ref{fig:noise_consistency}, TD is able to capture levels of perturbation severity much better and consistently than GS for many types of perturbations (e.g. Gaussian noise, frost, and elastic transform). This demonstrates that TD trend is more consistent with perturbation trend than GS, which further demonstrates the advantages of using features over pixels when computing topological properties.


\subsection{Comparison with IS}
 In our next experiment, we compare TD with IS on CelebA dataset where there are only face images and thus no distinct classes exist. We trained a GAN model (WGAN-GP~\cite{gulrajani_improved_2017}) on the training set of CelebA; original images were cropped to be of size 64$\times$64, and the model was then trained on them for 200 epochs with a batch size of 64. We recorded TD and IS along with the training process: every 4 epochs we fed the randomly sampled noise vector (remained fixed for different epochs) to the final model; we then computed TD and IS on the generated and real images.

\begin{figure}[tbh]
  \includegraphics[width=\linewidth]{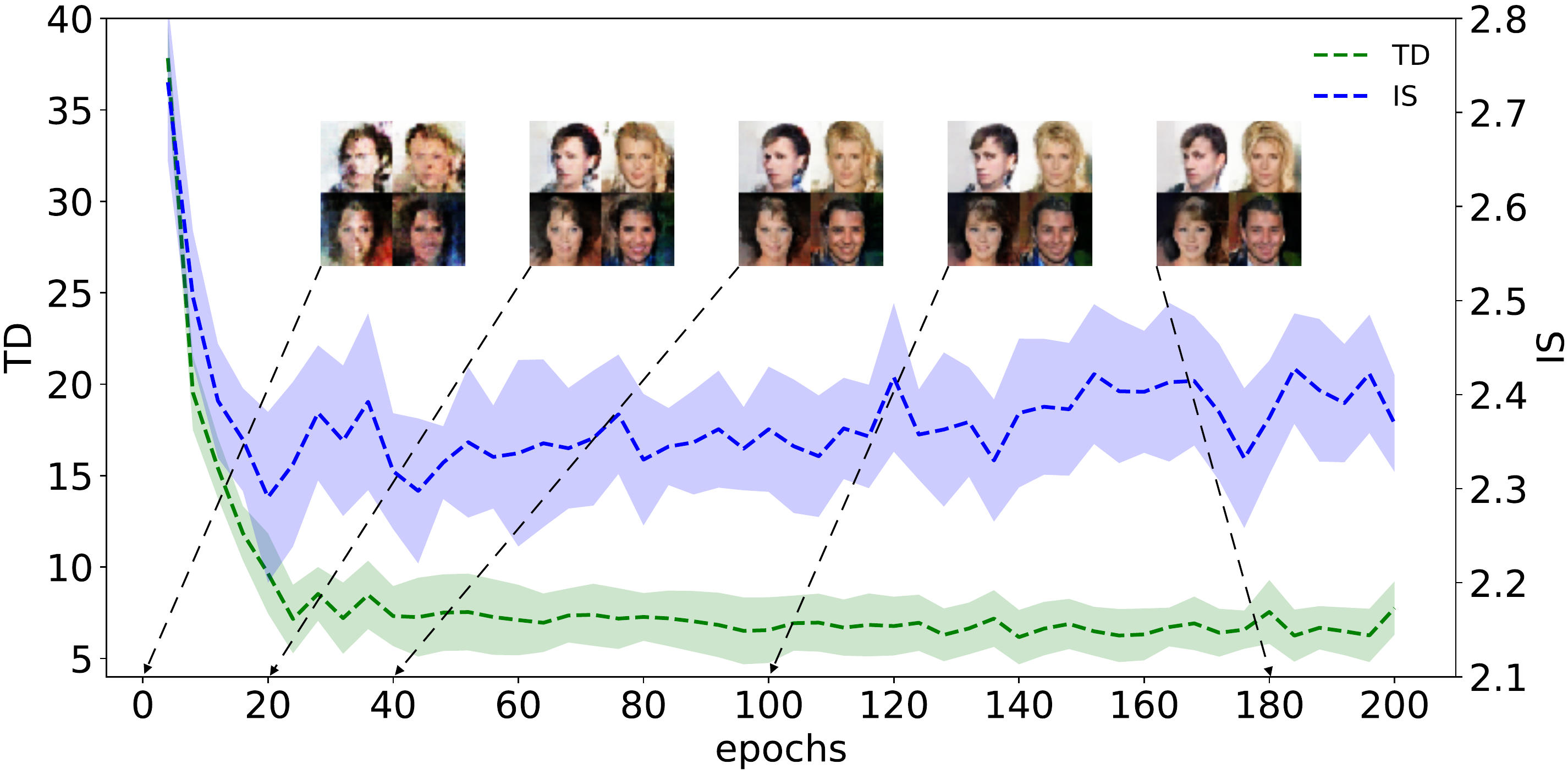}
  \caption{Comparison of TD and IS on generated images by WGAN-GP along with training process on the CelebA dataset. Results are averaged over 10 groups, each of which consists of 500 real images (randomly sampled from the original dataset) and generated images.}
  \label{fig:td_vs_is}
\end{figure}

Figure~\ref{fig:td_vs_is} shows the comparison results of TD and IS on the CelebA dataset. TD shows a great correspondence to the quality of generated images (i.e., decreasing trend with the improved quality of images). By contrast, IS fails to do so; at the early stages of training (before 20 epochs) it decreases as the quality of the generated images increases -- in contrast to what is expected -- and eventually loses its discrimination power at the remaining epochs. In summary, TD shows superiority over IS for evaluating the quality of images from datasets such as CelebA.


\subsection{Pixels vs. features}

Finally we performed an ablation study to compare the usage of pixels vs features when computing TD. We trained two WGAN-GP models, respectively, on Fashion-MNIST(trained for 100 epochs) and CIFAR10 (trained for 200 epochs) datasets. We then computed pixel-based and feature-based TD between images generated by WGAN-GP trained for different number of epochs and real images, randomly sampled from each dataset.

\begin{figure}[tb]
  \includegraphics[width=\linewidth]{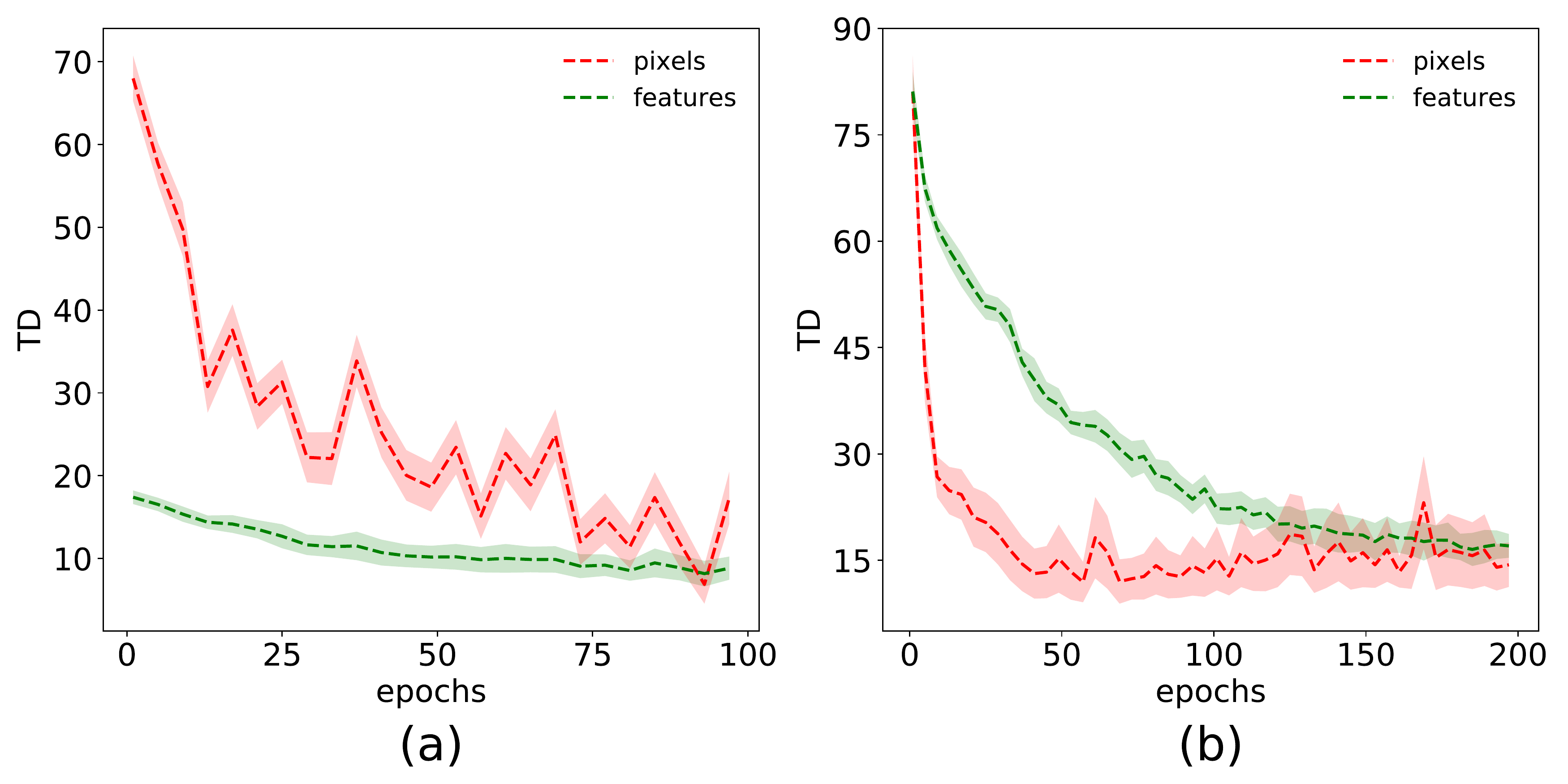}
  \caption{Comparison of TD based on pixels and features on different datasets. Results are averaged over 10 groups, each of which consists of 500 real images (randomly sampled from the original dataset) and generated images. (a) Fashion-MNIST. (b) CIFAR10. }
  \label{fig:pixels_vs_features}
\end{figure}

As can be seen clearly from Figure~\ref{fig:pixels_vs_features}, for both datasets, feature-based TD are able to demonstrate better performance in terms of discrimination and consistency. This attributes to the better generalisation of learned features than raw pixels, which is one of the most significant advances of deep neural networks~\cite{bengio_representation_2013}.

\section{Conclusion}
\label{Conclusion}

In this work, we introduced Topology Distance (TD), a novel metric to evaluate GANs by considering the topological structures of latent manifold of real and generated images. In a range of experiments we have compared TD with Inception Score (IS), Fr\'echet Inception Distance (FID), Kernel Inception Distance (KID), and Geometry Score (GS), and have demonstrated its advantages and superiority over them in terms of consistency with human judgement, as well as other quantitative measures of change in image quality. TD is capable of providing new insights for the evaluation of GANs, and it thus can be used in conjunction with other metrics when evaluating GANs.



\bibliography{ms}
\bibliographystyle{icml2020}

\renewcommand{\thesection}{\Alph{section}}
\renewcommand{\thefigure}{\Alph{figure}}

\onecolumn
\icmltitle{Supplementary Material for Topology Distance}

\vskip 0.3in

\setcounter{section}{0}
\section{Comparison of Sample Quality Correlation}

An essential property of a metric for evaluating GANs is its good correlation to the quality of generated samples. We thus performed a comprehensive comparison of image quality correlation, among Fr\'echet Inception Distance (FID), Kernel Inception Distance  (KID), Geometry Score (GS), Inception Score (IS) and our proposed Topology Distance (TD), on three datasets (i.e. Fashion-MNIST, CIFAR10 and CelebA). Specifically, for each dataset, we first trained a WGAN-GP model for 200 epochs with a batch size of 64. We then computed the scores of each metric between real images and generated images for every 4 epochs. Results were averaged over 10 groups, each of which consisted of 500 real images (randomly sampled from each original dataset) and corresponding generated images. The results are shown in Figure~\ref{fig:metrics_vs_epochs_multirun}. Compared with other metrics, TD demonstrates better (vs. GS and IS) or comparable (vs. FID and KID) performance of sample quality correlation. It is worth noting that the advantages of our proposed TD over other metrics (as we have shown in other experiments) make TD stand out as a robust and strong alternative to evaluate GANs.

\setcounter{figure}{0}
\begin{figure}[tbh]
  \includegraphics[width=140mm]{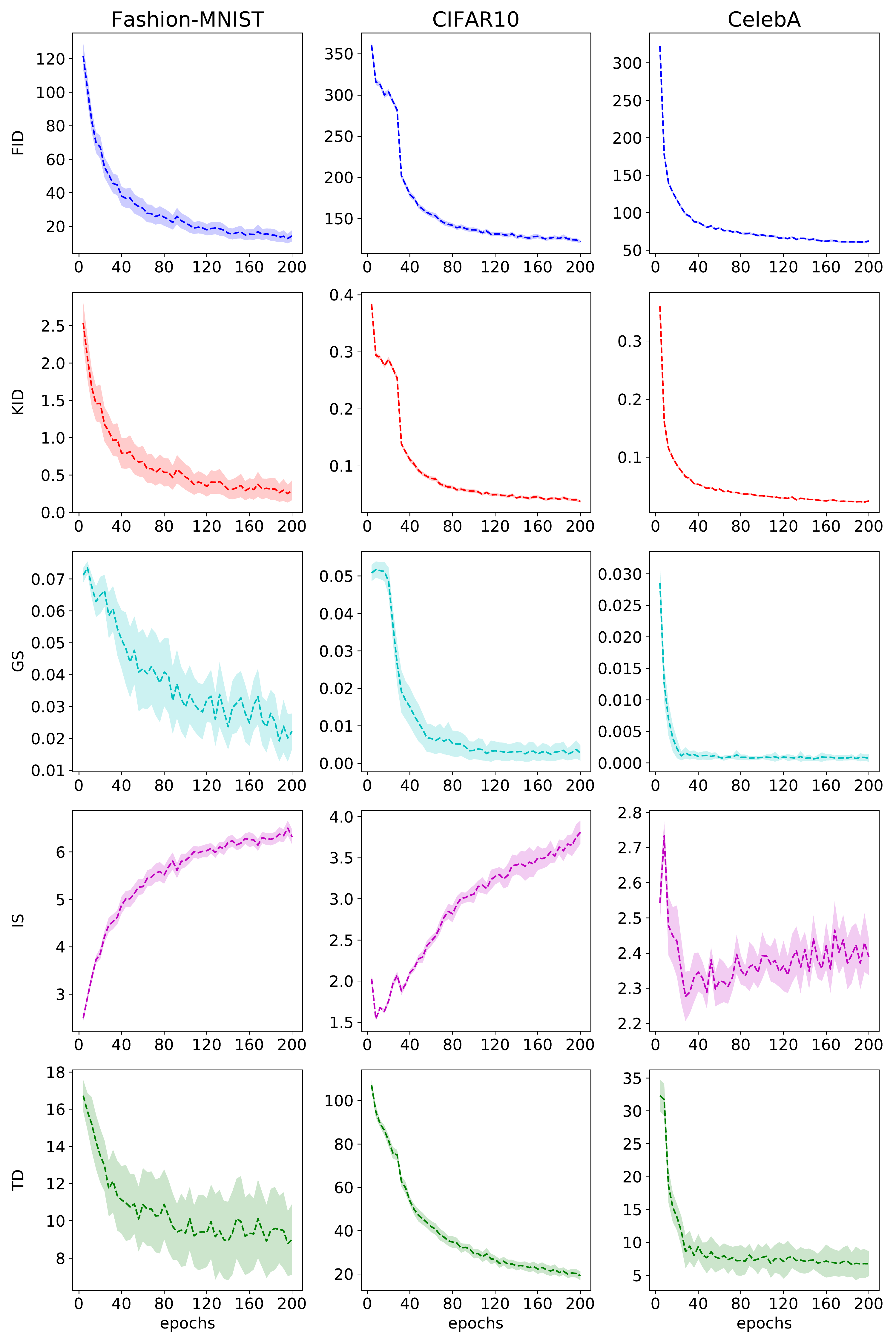}
  \centering
  \caption{Comparison of sample quality correlation among FID, KID, GS, IS and our proposed TD, on Fashion-MNIST, CIFAR10, and CelebA. Results are averaged over 10 groups, each of which consists of 500 real images (randomly sampled from the original dataset) and generated images.}
  \label{fig:metrics_vs_epochs_multirun}
\end{figure}

\end{document}